\documentclass[runningheads,a4paper]{template/llncs}
\authorrunning{Moulton, J, Karapetyan, N, Quattrini Li, A, Rekleitis, I}

\usepackage{amssymb}
\usepackage{graphicx}
\usepackage[tight,footnotesize]{subfigure}
\usepackage{amsmath, wrapfig, url} 
\usepackage{amssymb}  
\usepackage{hyphenat}
\usepackage[table]{xcolor}
\usepackage{algorithm}
\usepackage[noend]{algpseudocode}
\usepackage{float}
\usepackage{enumitem}
\long\def\invis#1{}
\usepackage{hyphenat}

\usepackage{cite}
\usepackage{url}
\usepackage{siunitx}
\usepackage{xcolor}
\usepackage{soul}
\renewcommand{\figurename}{Fig.}

\newcommand\etal{\textit{et al.\ }}
\DeclareMathOperator{\atantwo}{atan2}

\newcommand\acomment[1]{\textcolor{red}{A:#1}}

\urldef{\mailsa}\path| {moulton, nare}@email.sc.edu, alberto.quattrini.li@dartmouth.edu|
\urldef{\mailsb}\path| yiannisr@cse.sc.edu|

\long\def\invis#1{}

\makeatletter
\def\BState{\State\hskip-\ALG@thistlm}
\makeatother

\begin{document}
\sisetup{quotient-mode=fraction}

\mainmatter  

\title{External Force Field Modeling for Autonomous Surface Vehicles\vspace{-0.1in}}


%
%
\author{Jason Moulton$^1$, Nare Karapetyan$^1$, Alberto Quattrini Li$^2$, and Ioannis Rekleitis$^1$\\
}

\institute{$^1$Computer Science \& Engineering Department, University of South Carolina,\\ $^2$Computer Science Department, Dartmouth College\\
\mailsa,\\ \mailsb\vspace{-0.2in}
}

%
%

\toctitle{Lecture Notes in Computer Science}
\tocauthor{Authors' Instructions}
\maketitle
\begin{abstract}
Operating in the presence of strong adverse forces is a particularly challenging problem in field robotics. In most robotic operations where the robot is not firmly grounded, such as aerial, surface, and underwater, minimal external forces are assumed as the standard operating procedures. The first action for operating in the presence of non\hyp trivial forces is modeling the forces and their effect on the robots motion. In this work an Autonomous Surface Vehicle (ASV), operating on lakes and rivers with varying winds and currents, collects wind and current measurements with an inexpensive custom-made sensor suite setup, and generates a model of the force field. The modeling process takes into account depth, wind, and current measurements along with the ASVs trajectory from GPS. \invis{and IMU data, together with the morphology and bathymetry of the environment.} In this work, we propose a method for an ASV to build an environmental force map by integrating in a Gaussian Process the wind, depth, and current measurements gathered at the surface. We run extensive experimental field trials for our approach on real Jetyak ASVs. Experimental results from different locations validate the proposed modeling approach.
\end{abstract}
\section{Introduction}
\begin{wrapfigure}[13]{r}{0.5\textwidth}
\centering
\vspace{-0.3in}
\includegraphics[width=0.5\textwidth]{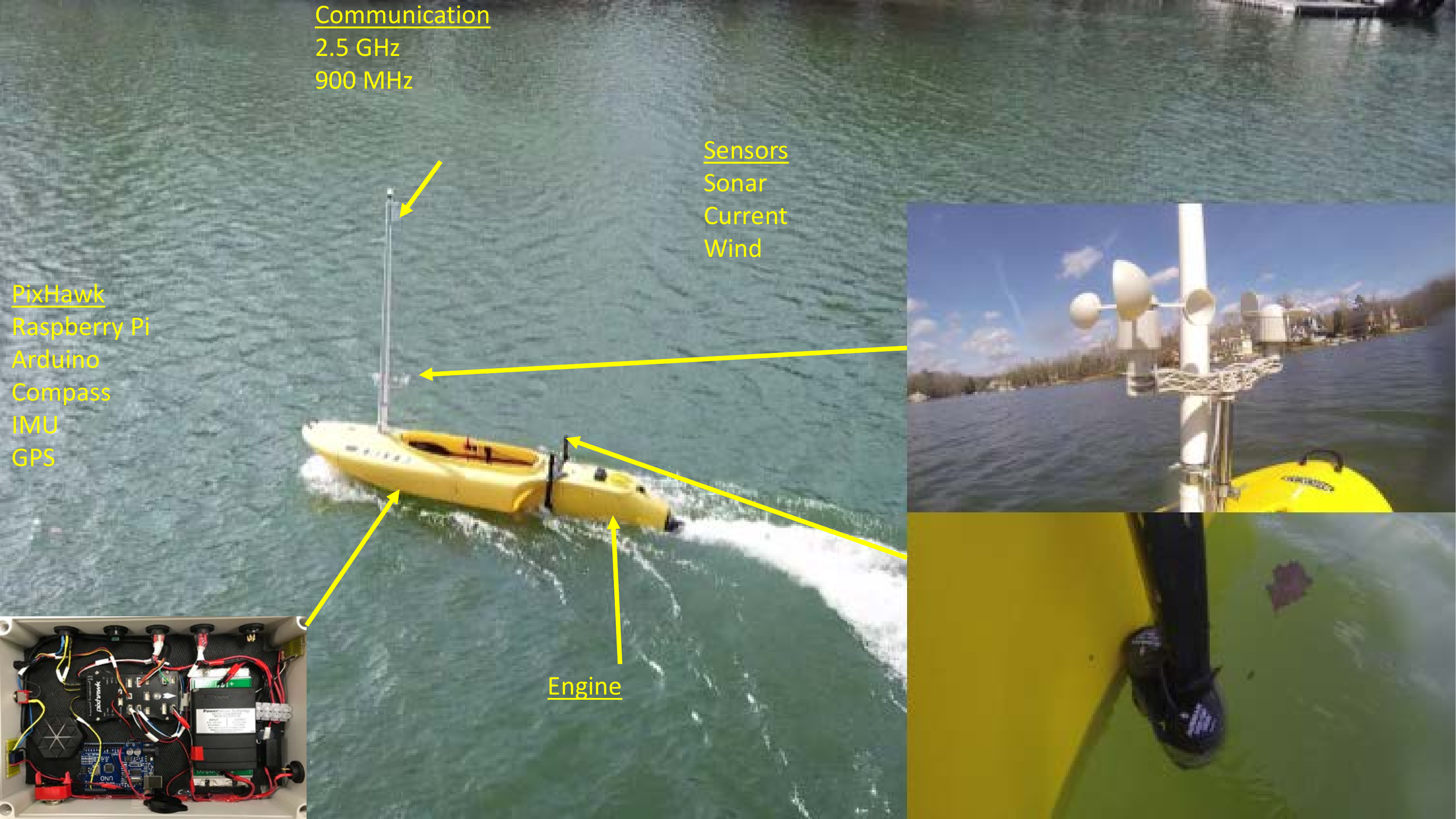}\vspace{-0.05in}
\caption{\label{fig:jetyak_2}UofSC autonomous surface vehicle outfitted with anemometer, depth sonar, and current sensors. }
\end{wrapfigure}

While robots become increasingly common, from the robotic vacuum cleaner and warehouse product\hyp moving robots, to the prospect of autonomous cars, robots are assumed to operate largely undisturbed. Most fielded robots operate on fairly flat grounds, with minimal wind, wave, and current forces. Contrary to these scenarios, we consider robots that are best suited to operate in environments restrictive to humans.  As such, capabilities to operate in unknown/dynamic environments, and in the presence of adverse external forces, are required to ensure that robots become ubiquitous and safe in many applications -- such as safe inspection of infrastructure~\cite{murphy2011use}, search and rescue~\cite{matos2016multiple}, environmental sampling~\cite{RekleitisICRA2018b}, monitoring of water quality~\cite{RekleitisISERMulti2016}, and mapping inaccessible regions in more efficient, less cost prohibitive means. 

In this work, using an Autonomous Surface Vehicle (ASV) -- see Figure \ref{fig:jetyak_2} -- we provide the following contributions: a reliable inexpensive platform for collecting  depth, wind, and current data in different environments and conditions; and a data processing and model derivation approach for spatially varying environments.
\vspace{-0.1in}

\begin{wrapfigure}[14]{r}{0.41\textwidth}
\centering
\vspace{-0.3in}\includegraphics[width=0.41\textwidth]{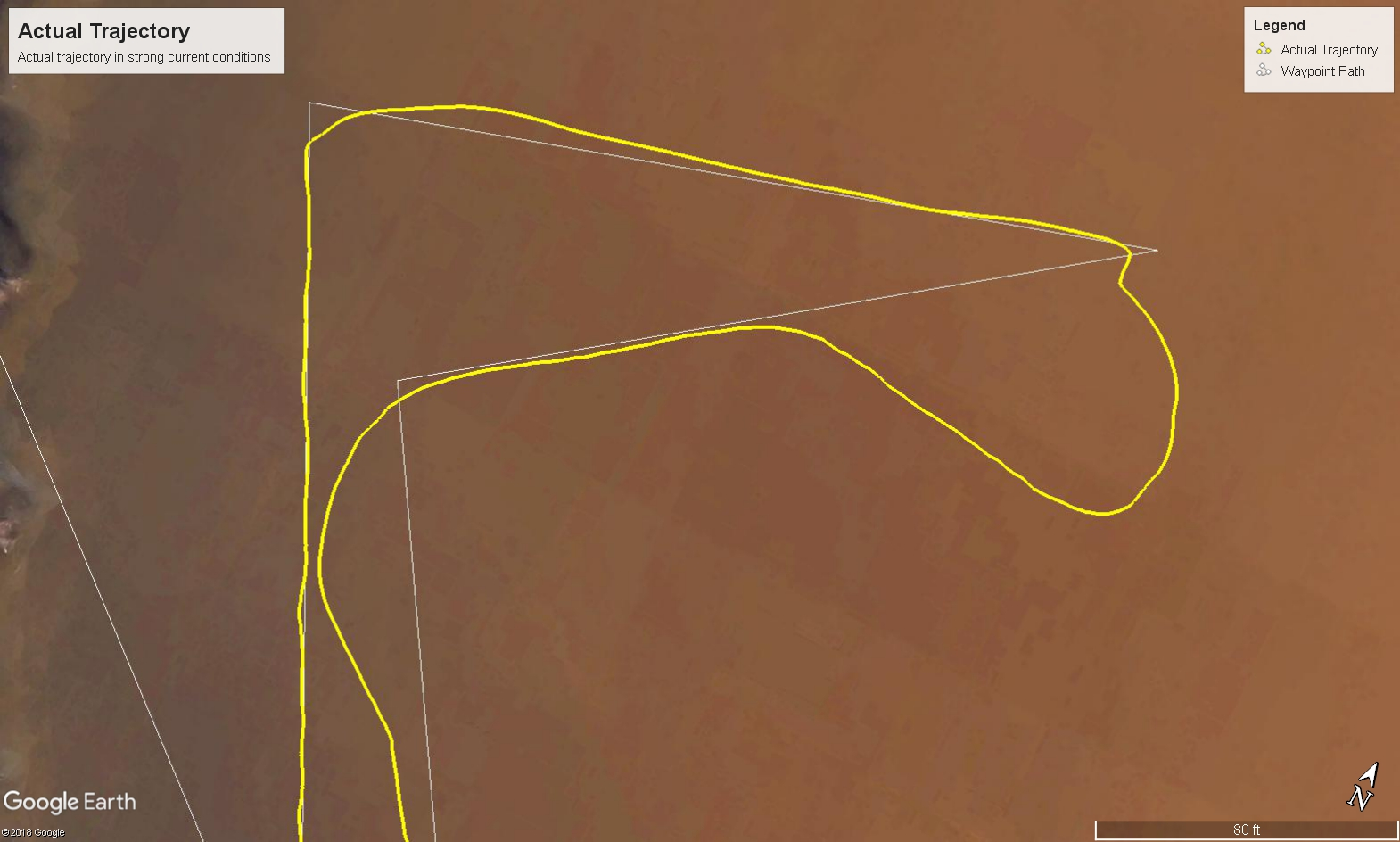}
\caption{\label{fig:problem_pic}ASV unable to maintain course in heavy current when turning in downstream directions. The white line represents ideal path and the yellow line the actual (GPS) trajectory of the ASV.}
\end{wrapfigure}
\subsection{Motivation}
\begin{figure*}[t]
	\centering
	\leavevmode
	\begin{tabular}{ccc}
	\subfigure[\label{fig:ideal_n-s}] {\includegraphics[width=0.32\textwidth]{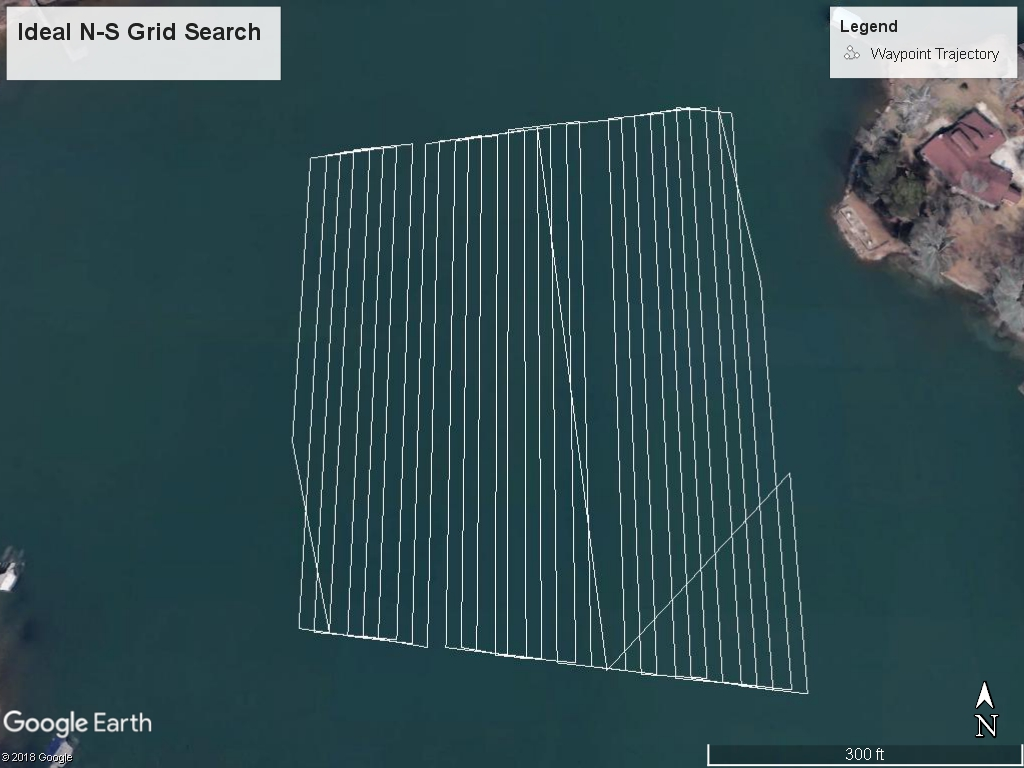}\label{fig:s1}}&
	\subfigure[\label{fig:boat1_calm_wind}] {\includegraphics[width=0.32\textwidth]{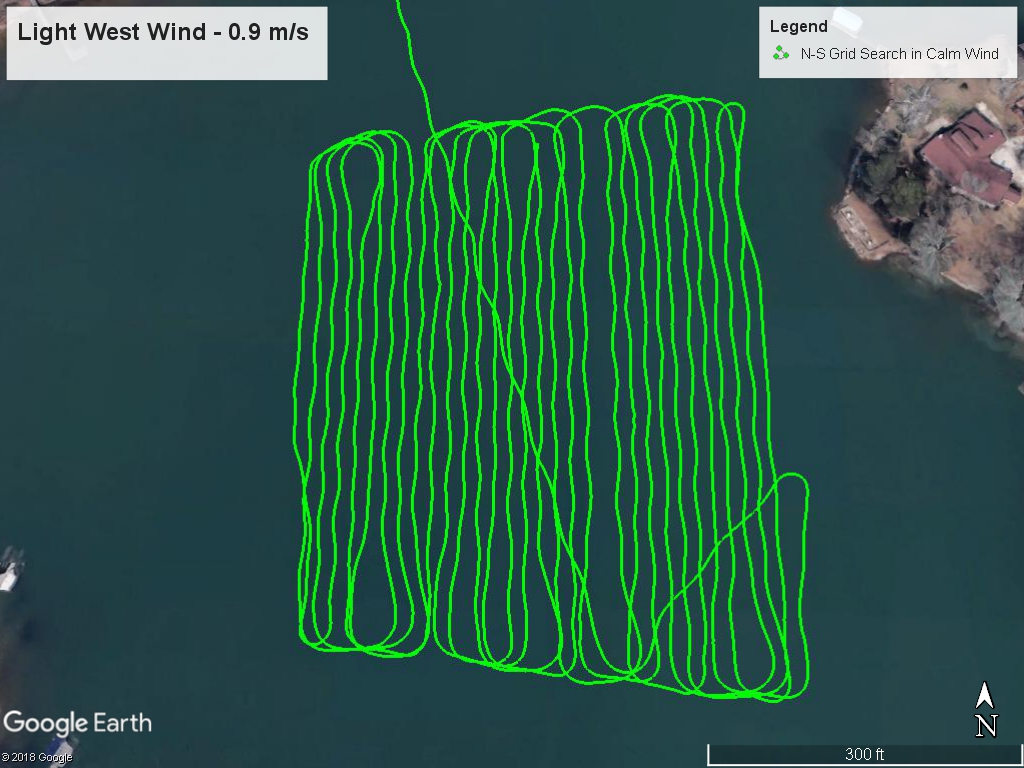}\label{fig:s2}}&
	\subfigure[\label{fig:grid_search_windy}]{\includegraphics[width=0.32\textwidth]{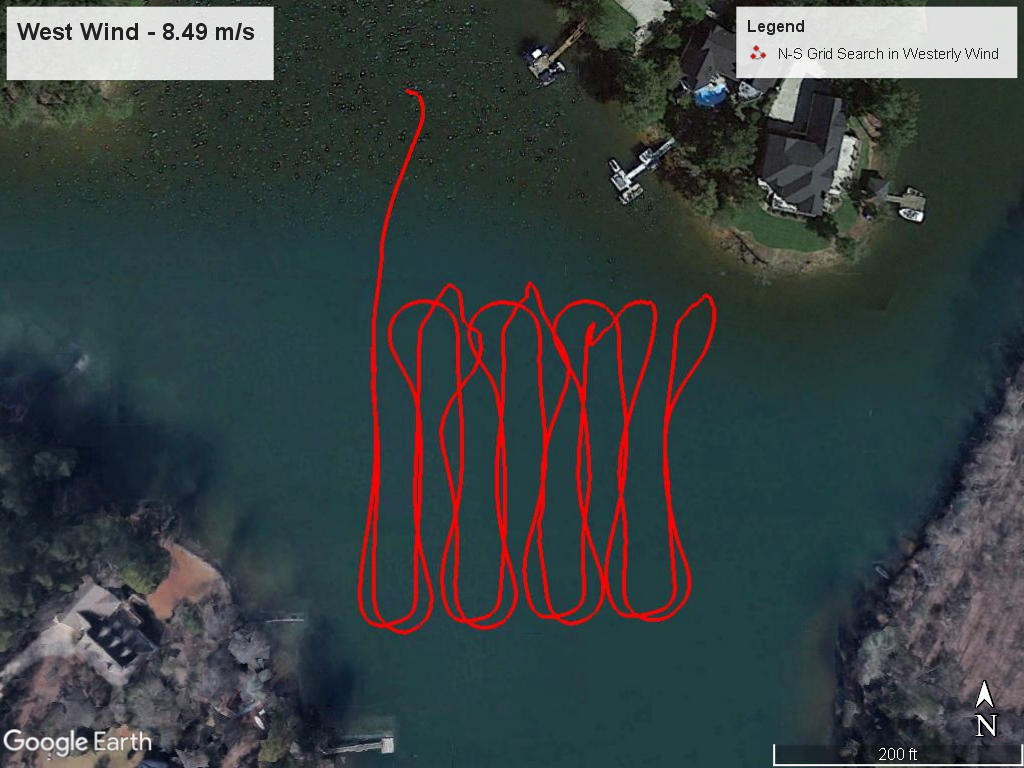}\label{fig:s3}}\vspace{-0.15in}
	\end{tabular}
	\caption{Ideal North-South \subref{fig:s1} grid search missions are used to provide a baseline for measuring tracking performance.  Trajectories for grid searches conducted in \subref{fig:s2} calm conditions and \subref{fig:s3} a subset in 8.49 m/s wind illustrating the effects of the wind on the ASV. 
	\label{fig:wind_behaviors}}
\end{figure*}

Currently, as observed in a variety of experiments -- see Figures \ref{fig:problem_pic} and \ref{fig:wind_behaviors} for some examples -- an ASV relying on a conventional robotic actuator \invis{proportional, integral, derivative (PID)} controllers for way\hyp point navigation is unable to maintain its course when faced with non trivial external forces such as heavy wind or current.  Due to the PID controller being tuned for conditions which are stable, there is no simple method to provide tuned coefficients for dynamic environments where wind and currents are always changing. This drives our motivation to pro\hyp actively model, plan for, and adapt to these dynamics so the robot can maintain its course and not miss large swaths of its planned trajectory.

\invis{\subsection{Problem Statement}}
\invis{
Operating in marine environments that can have highly dynamic forces acting on an Autonomous Surface Vehicle (ASV), it is highly beneficial to measure, model, and predict the behavior of these forces, in order to enable intuitive route planning and prolonged deployments where initially, conditions may have been assessed to be outside the robots operating limitations.   In this work, we collect and model the environmental variables directly impacting an ASV in rivers and lakes. \acomment{Repetition, we can streamline the whole section avoiding repetitions.} The variables we strive to model and forecast are wind and current (which depends on depth), both of which present specific challenges to a robot operating in such environment.  
}

\subsection{Related Work}
Current research modeling wind and currents mainly focuses on large scales, with applications to oceanic navigation and power generation. In the problem addressed in this work, the effects of environmental forces impacting the platform are greater considering the size of the ASV.
For instance the work of Soman \etal \cite{soman2010review} reviewed existing wind prediction strategies for optimizing efficiency and profits in power generation applications.  Their work focuses both on long term and short term forecasting, but mostly as it pertains to direction, since the turbines they are designing for are stationary.  
One more closely related study is by Al-Sabban \etal \cite{al2013wind}, which focuses on the effect of wind on Unmanned Aerial System (UAS). They implement a hybrid Gaussian distribution of a wind field and a slightly modified Markov Decision Process (MDP) to identify the optimal path and optimal power consumption trajectory for a UAS. This work may prove valuable in our future work, but at the current time, we are concerned with an environment with leeward and windward effects caused by surrounding landscape. 
Another work, by  Encarna{\c{c}}ao and Pascoal \cite{encarnaccao2001combined}, studied the  problem of developing control systems for marine crafts that are able to follow trajectories to track another boat, under the effects of water currents. Their models center on inertial tracking and compensation of roll, pitch, and yaw rates after the force is sensed to provide course corrections. In our work, we aim at being more proactive, in that we actively measure and model environmental variables with the end\hyp state intent on taking corrective control measures prior to the robot coming under the effects of the wind or currents. Finally, Hsieh \etal~\cite{hsieh2012robotic} have provided many contributions related to the problem of mapping the effects current phenomena with their design of a control strategy for collaborative underwater robots to track coherent structures and manifolds on generally static conservative flows~\cite{michini2014robotic,kularatne2018optimal,hsieh2018small}. Huynh \etal presented a path planning method for minimizing the energy consumption of an autonomous underwater vehicle (AUV) \cite{Huynh2015}. Their work addresses varying ocean disturbances that are assumed to not exceed the capabilities of the AUV. These works, while addressing aspects of collecting and modeling dynamic environmental characteristics, do not address fully the same time and space considerations that impact our lightweight, small\hyp scale ASV. To establish models that can drive future adaptive controls and online planning strategies for our ASV, we adopt the following technical approach.
\section{System and Methodology}

In this section, we present the proposed hardware setup, calibration, and verification of inexpensive sensors in a controlled environment, experimental environments, and data collection and processing required to develop comprehensive models of external forces. We break this into two components, the physical and technical characteristics associated with the design and build of our ASV, and the data gathering/processing approach we take to model depth, current, and wind. The modeling process takes into account wind and current measurements, the ASVs trajectory from GPS, compass, and IMU data, together with the morphology and bathymetry of the environment.

\invis{\subsection{Hardware Setup}}
We deployed our methods on a Mokai Es-Kape  platform\footnote{\url{http://www.mokai.com/mokai-es-kape/}} termed \emph{Jetyak}, based on the WHOI Jetyak~\cite{kimball2014whoi} ASV and conducted extensive calibration of the operational parameters in benign conditions. Furthermore, we performed several experiments to collect data under a variety of conditions. Such data has been used  to perform predictions to evaluate different machine learning techniques.
 In the following a brief overview of the platform is presented -- see our work \cite{RekleitisOceans2018b} for more details on our version of the jetyak.
The baseline system, capable of way\hyp point navigation and teleoperated control, consists of an ArduPilot \cite{ardupilot} PixHawk microcontroller utilizing IMU, GPS, and telemetry data. The sensors employed are intentionally inexpensive to enable their adoption by a large number of end\hyp users.  For instance, our anemometer\footnote{\url{https://learn.sparkfun.com/tutorials/weather-meter-hookup-guide}} is readily available for building and programming as a kit. It provides wind speed and direction updates at \SI{4}{\Hz} costing about 100 US dollars, whereas a state\hyp of\hyp the\hyp art ultrasonic anemometer with user accessible serial output costs over 1,500 US dollars. We employ the same constraint for the depth sonar and water current sensors, with none of the sensors costing more than 200 US dollars. Selected water current sensors\footnote{\url{http://www.raymarine.com/view/index.cfm?id=555}} are simple hall effect sensors. The  depth sensor\footnote{\url{http://www.cruzpro.co.nz/active.html}} mounted on the ASV is National Marine Electronics Association (NMEA) 0183 compliant\footnote{\url{https://www.nmea.org/content/nmea_standards/nmea_0183_v_410.asp}} and outputs serial text to allow Robot Operating System (ROS) \cite{quigley2009ros} based software direct access to the measurements. In addition to \SI{2.4}{\GHz} remote control and \SI{900}{\MHz} remote frequency communications, we employ a Raspberry Pi to host the ROS framework recording all telemetry, depth, current, and wind data collected. \invis{For more information on the utilized design, see our work  Moulton \etal\cite{RekleitisOceans2018b}.}

\begin{wrapfigure}[15]{r}{0.5\textwidth}
\centering
\vspace{-0.3in}
\includegraphics[width=0.4\textwidth, trim={.5cm 0 .5cm 0},clip]{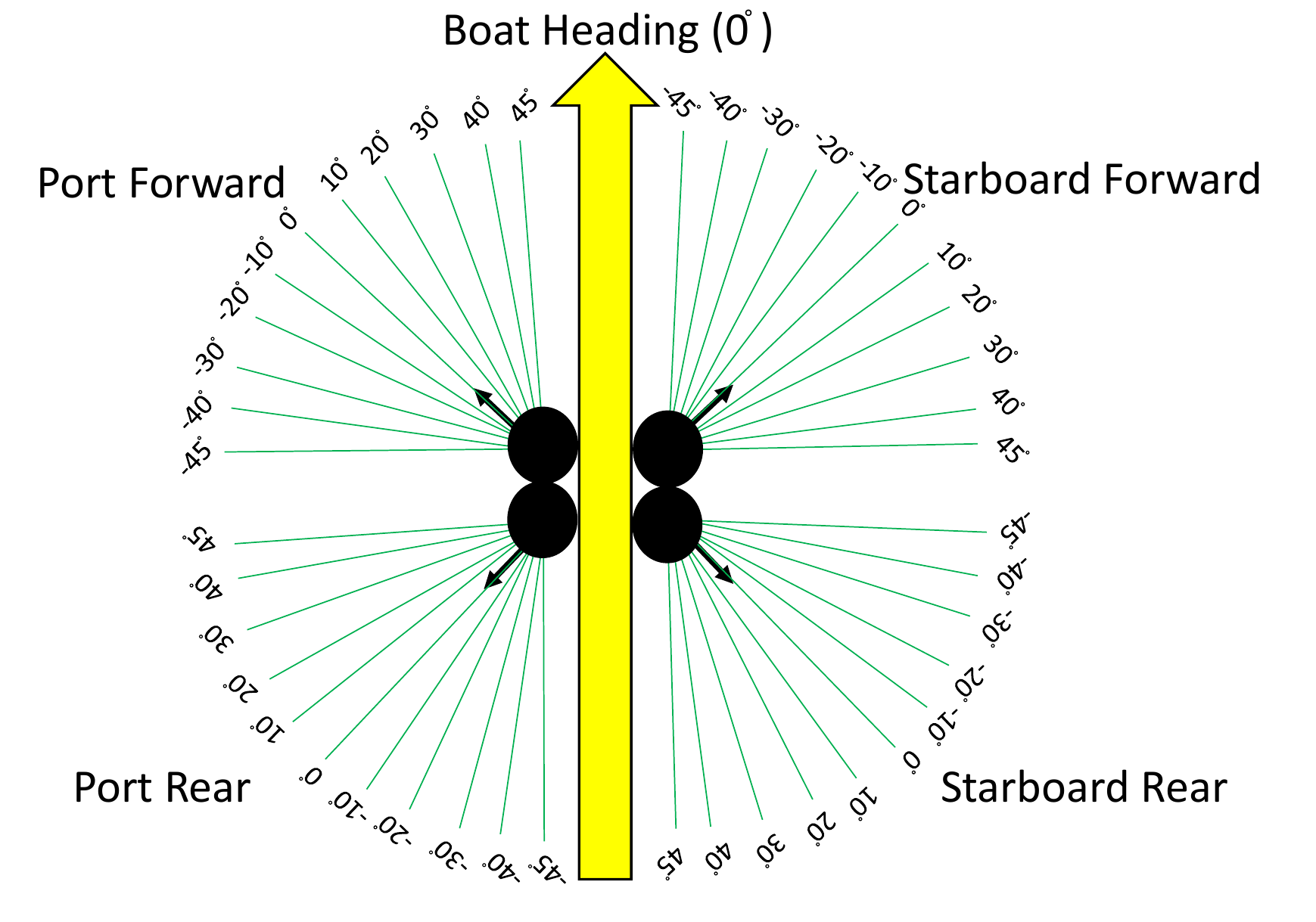}
\caption{\label{fig:sensor_coverage}Mounting configuration for surface current sensors allows selection of the highest sensor reading for F1 and its next highest neighbor for F2. Calculations for speeds and directions are explained below.}
\end{wrapfigure}
Mounting inexpensive, off\hyp the\hyp shelf, sensors requires experimentation both on the mounting position and orientation, as well as reverse engineering to build and calibrate the micro\hyp controller servicing the sensor.  Being a 1-d sensor, the depth sonar requires the smallest effort, since it returns depth readings directly to the USB port of the Raspberry Pi.  The only challenge with mounting a depth sensor is ensuring it remains submerged and avoids cavitation interference resulting from the physical properties of the ASV's hull. The same challenge applies to selecting the location and orientation of the current sensors.  Cavitation, air pockets, and water deflection from the hull and the other sensors have an adverse impact on the water flow through the sensor. The least challenging sensor to mount is the anemometer, simply requiring a stand-off bracket to attach it to the mast. Prior to installation, calibration of the current, wind, and depth sensors is performed to ensure the accuracy of the recorded data. A full tutorial on building the discussed platform with all diagrams and sensor driver code is available on the UofSC Autonomous Field Robotics Lab (AFRL) website \cite{doc}, including lessons learned for programming inexpensive micro\hyp controllers and placement/alignment of sensors.

\invis{ using a hand-held anemometer and known depth readings to compare to our implementation, respectively.  In particular, calibrating the current sensors requires the use of a controlled water flow setup }

The facilities of the Hydraulics Laboratory at  Civil and Environmental  Engineering department were used to calibrate the sensors and establish ground truth estimates. Throughout all calibrations, the PixHawk with ArduPilot code, ROS node, Arduino microcontroller programs are run to ensure proper operation when added to the ASV. Laboratory experiments using controlled flow demonstrated that the sensors are effective for water current component measurement at a $\pm 45^\circ$ orientation to the sensor. As such, different setups have been tested to achieve the best coverage. After several attempts to provide accurate current readings covering all directions, we arrived at the proposed solution of four current sensors offset from each other at $90^\circ$ and offset from the ASV's reference frame by $45^\circ$; see Figure \ref{fig:sensor_coverage} for the proposed configuration. Using the highest measurement and highest neighbor measurement, we are able to accurately calculate the current in the boat's reference frame and transform it to the world reference frame by subtracting the boat's x and y velocity components. 
The high-level calculations are described in the following section.

\subsection{Data Gathering and Processing}

The wind and current sensors record data are influenced by the motion of the ASV. More specifically, the sensor measurement ($\overrightarrow{R}$) is the vector sum between the motion of the ASV ($\overrightarrow{A}$) and the real value of the physical phenomenon ($\overrightarrow{W}$). The motion of the ASV can be inferred by the GPS velocity and compass sensors. Therefore, the true value can be estimated as $\overrightarrow{W}=\overrightarrow{R}-\overrightarrow{A}$.

The current sensors are mounted in fixed locations and measure scalar current velocities ($f$).  We select the component forces for ($\overrightarrow{F}$) by assigning the highest measured force to $F1$ and the highest of $F1's$ two neighboring sensor readings to $F2$.  The forces are then offset by $45^\circ$ to account for the angular velocities read by the sensor as the ASV ($\overrightarrow{A}$) traverses the current ($\overrightarrow{C}$).  In our optimal setup, we have aligned four sensors, one offset at $45^\circ$ on each quadrant of the boat, illustrated in Figure \ref{fig:sensor_coverage}. This provides the equation $\overrightarrow{C}=\overrightarrow{F}-\overrightarrow{A}$ which we will solve using the trigonometric properties resulting from our sensor array alignment.

Once accurately aligned, the collected data were combined using a Gaussian Process (GP) mapping technique~\cite{rasmussenGP} to build a model of forces and depth. Initial modeling and correlation uses boat heading, GPS velocity of the boat, depth measurements, wind speed and direction measurements, and four current measurements as input.
More formally, given a phenomenon $f(\mathbf{x})$, a GP can be used to estimate $f(\mathbf{W})$ at locations $\mathbf{W}=[\mathbf{w}^1, \mathbf{w}^2, \ldots \mathbf{w}^k]$ with a posterior distribution fitted over noisy measurements $\mathbf{Y} = [y^1, y^2, \ldots, y^n]$ collected by the robots at the corresponding GPS locations $\mathbf{X} = [\mathbf{x}^1, \mathbf{x}^2, \ldots, \mathbf{x}^n]$:
\begin{equation}
p(f(\mathbf{W})\mid \mathbf{W}\mathbf{X}, \mathbf{Y}) \sim \mathcal{N}(\mathbf{\mu}_{\mathbf{W}}, \mathbf{\Sigma}_{\mathbf{W}}).
\end{equation}
\begin{wrapfigure}[13]{r}{0.5\textwidth}
\centering
\includegraphics[width=0.6\textwidth, trim={2cm 0 0 2.5cm},clip]{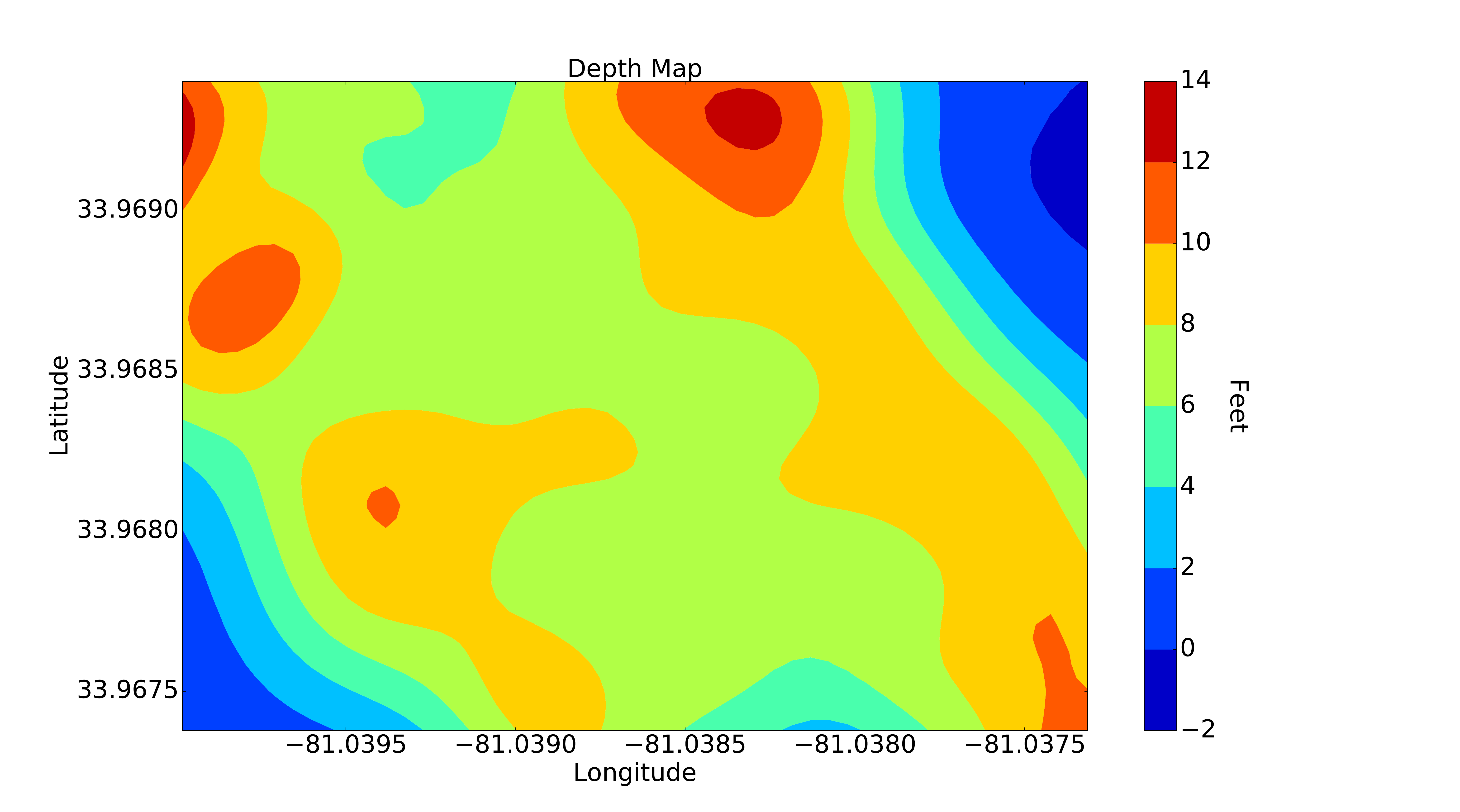}
\caption{\label{fig:depth_map}Depth map of Congaree River bottom resulting from GP.}
\end{wrapfigure}
As typically done in the mainstream approach, assuming a zero-mean GP, the estimate of the phenomenon is given by the mean vector $\mathbf{\mu}_{\mathbf{W}} = K(\mathbf{W},\mathbf{X}) \text{cov}(\mathbf{Y})^{-1}\mathbf{Y}$, where $\text{cov}(\mathbf{Y}) =  K(\mathbf{X}, \mathbf{X}) + \sigma_n^2 I_{q}$ is the correlation between observed values and $\sigma_n^2$ is the noise affecting the measurements $\mathbf{Y}$.
The covariance matrix  is calculated as $\mathbf{\Sigma}_{\mathbf{W}} = K(\mathbf{W},\mathbf{W}) - K(\mathbf{W},\mathbf{X}) \text{cov}(\mathbf{Y})^{-1} K(\mathbf{W},\mathbf{X})^T$. 
For accuracy evaluation, we tested different \textbf{\textit{K}}() kernels. Kernels used for this initial comparison include linear, ExpQuad, Matern $3/2$, and radial basis function (RBF), among which best performance was achieved by Matern $3/2$, expressed by the following equation:
\begin{equation}
k(\mathbf{x}^i,\mathbf{x}^j|\Theta ) = \sigma_n^2(1+\frac{\sqrt3r}{\sigma_l})exp(-\frac{\sqrt3r}{\sigma _l}),
\end{equation}

\noindent where $r$ is the Euclidean distance between $\mathbf{x}^i$ and $\mathbf{x}^j$, and $\sigma_l$ is a positive parameter.
Using the observations $\mathbf{X}$ and $\mathbf{Y}$ through the optimization of hyperparameters of the GP, predictions can be obtained.

\vspace{-0.1in}
\section{Experiments}
\begin{figure*}[t]
		\centering
		\leavevmode
		\begin{tabular}{cc}
		\subfigure[\label{fig:river_parallel}] {\includegraphics[width=0.5\textwidth]{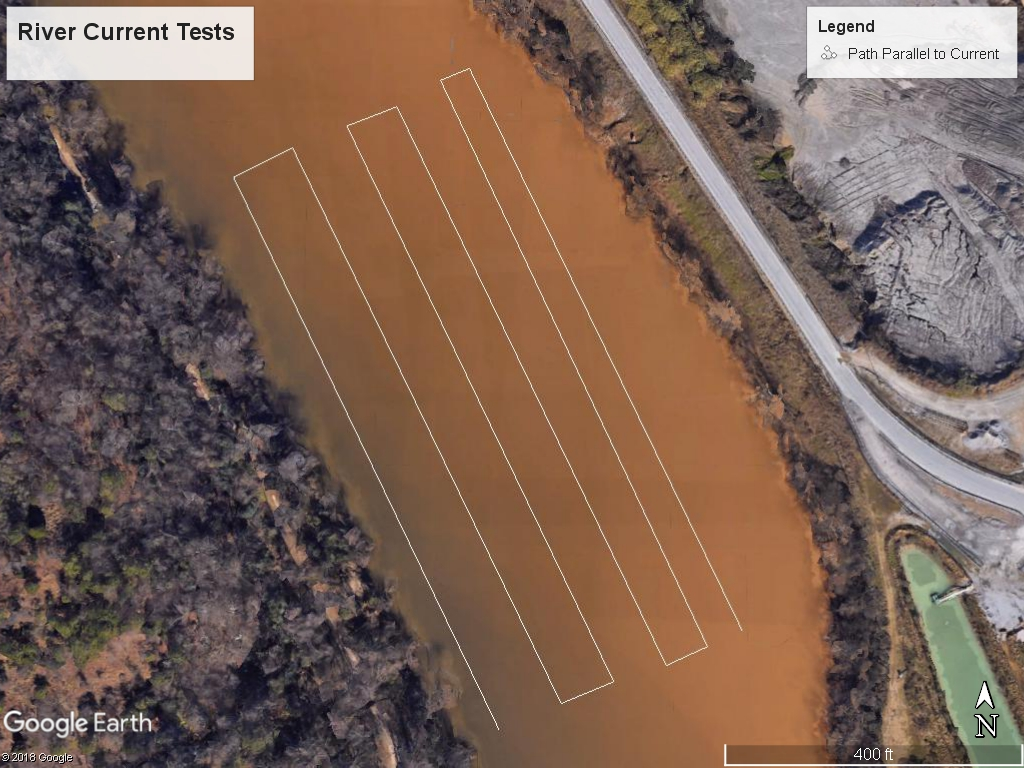}}&
		\subfigure[\label{fig:river_perp}] {\includegraphics[width=0.5\textwidth]{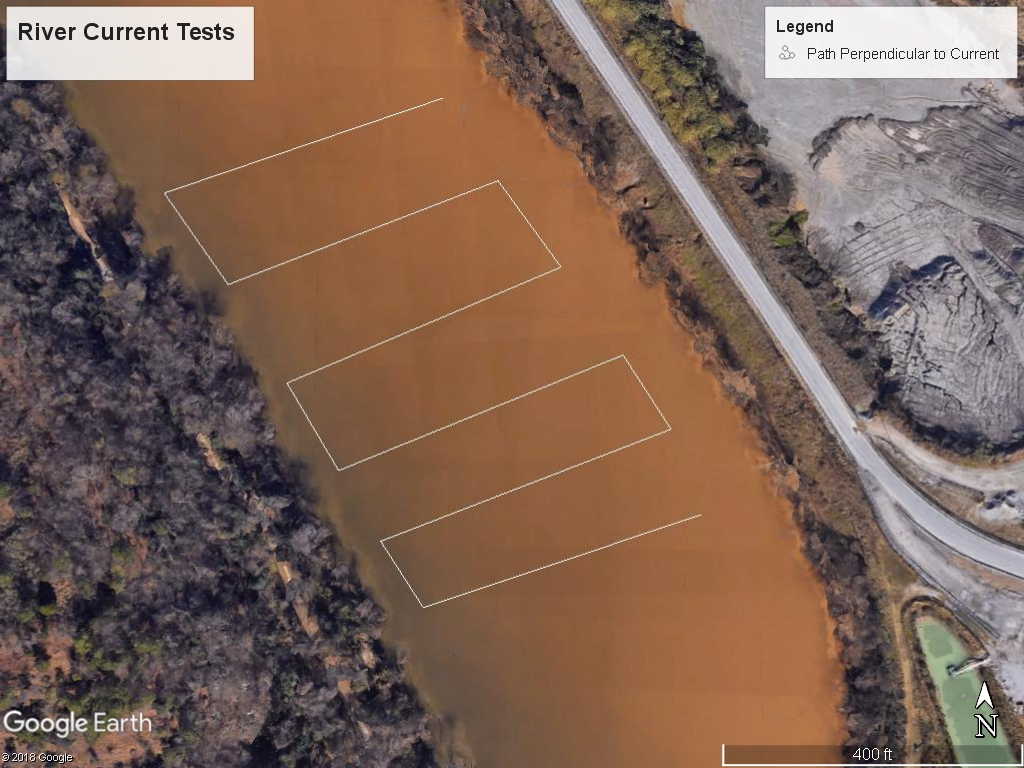}}\vspace{-0.15in}
		\end{tabular}
			\caption{Baseline testing pattern parallel\subref{fig:river_parallel} to the predominant current and perpendicular \subref{fig:river_perp} to the predominant current on the Congaree River, SC. 
    \label{fig:maps}}
	\end{figure*}

Experiments were carried out with a Jetyak equipped with the sensor suite described above in two different environments: a lake (Figure \ref{fig:wind_behaviors}) in a \SI{100 x 100}{\m} region, relatively calm; and a river (Figure \ref{fig:maps}), where the conditions are changing over time depending on rain and planned discharges by the local hydro-electric company. Our methodology uses standard grid\hyp search patterns to establish a baseline for performance comparison.\invis{single or multiple robot grid searches \cite{icra2017}.} \invis{Wind, current, and depth data were collected in Lake Murray, SC, USA. In addition to this baseline for a stable environment, we collected data in the challenging, fast moving and ever-changing Congaree River for a dynamic environment.} 

We then process the collected data in two stages.  First, we verify all time stamp and world orientation data by converting the PixHawk data logs to keyhole markup language (KML) for visual inspection of the missions, sequences and trajectories of the robots during the field trial.  Once time stamps are verified to contain no gaps, we then process the ROS bag file by a Python script to align time stamps from each sensor, based on an approximate time synchronizer scheme\footnote{\url{http://wiki.ros.org/message_filters/ApproximateTime}}. 
\invis{which organizes the data fields we are extracting and calls back Python's internal ApproximateTimeSynchronizer function.} 
It is at this point where we can verify the integrity of the data collected. Once complete, we are able to deem our system, scripts, calculations, and data as sound for further processing. The second stage processes the data for transformations, visualizations, and GP predictions as described in the previous section.

\begin{figure*}[b]
		\centering
		\leavevmode
        \begin{tabular}{cc}
		\subfigure[\label{fig:wind_speed_pred}] {\includegraphics[width=0.54\textwidth, trim={2cm 0 0 2.5cm},clip]{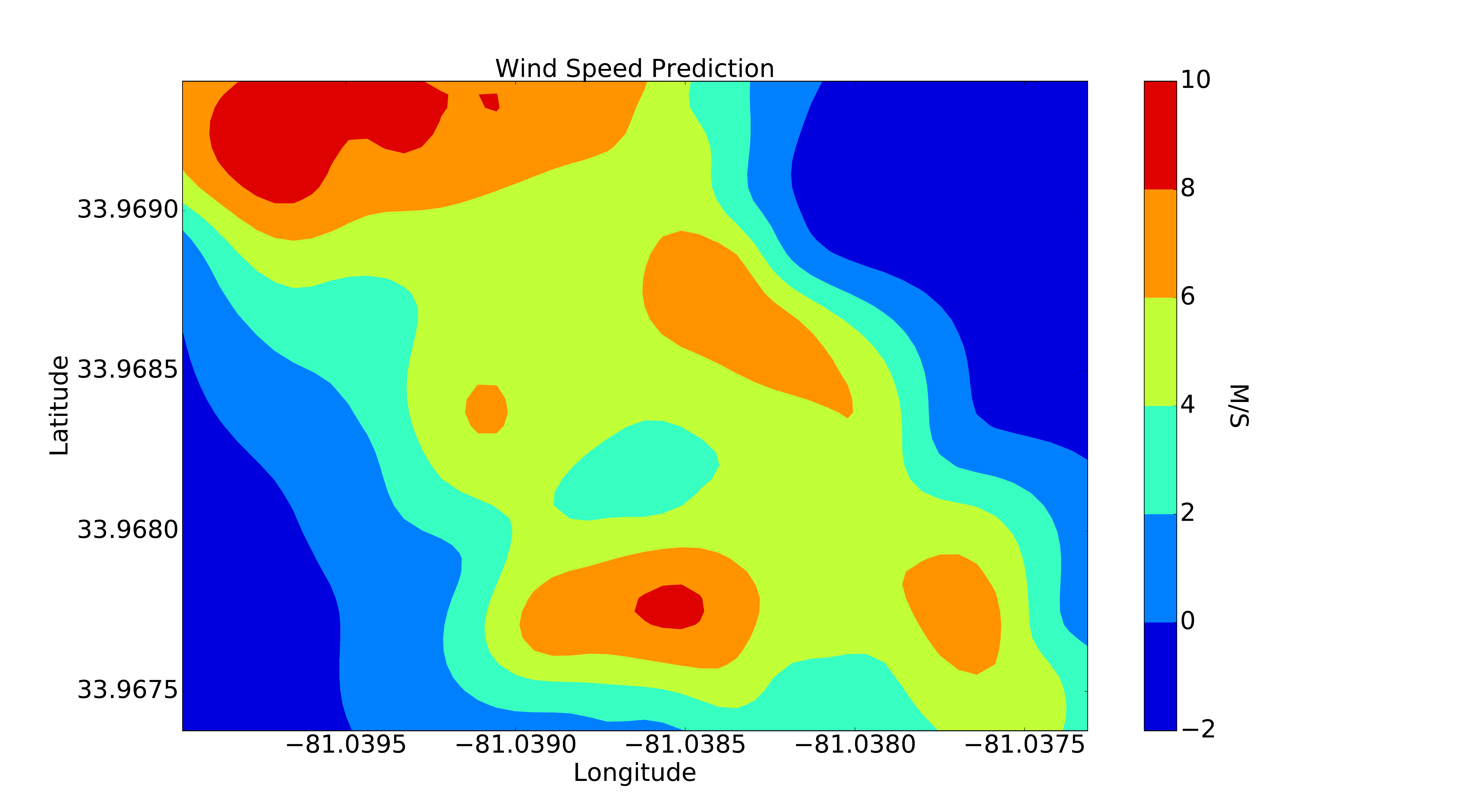}}&
			\subfigure[\label{fig:wind_dir_pred}] {\includegraphics[width=0.54\textwidth, trim={2cm 0 0 2.5cm},clip]{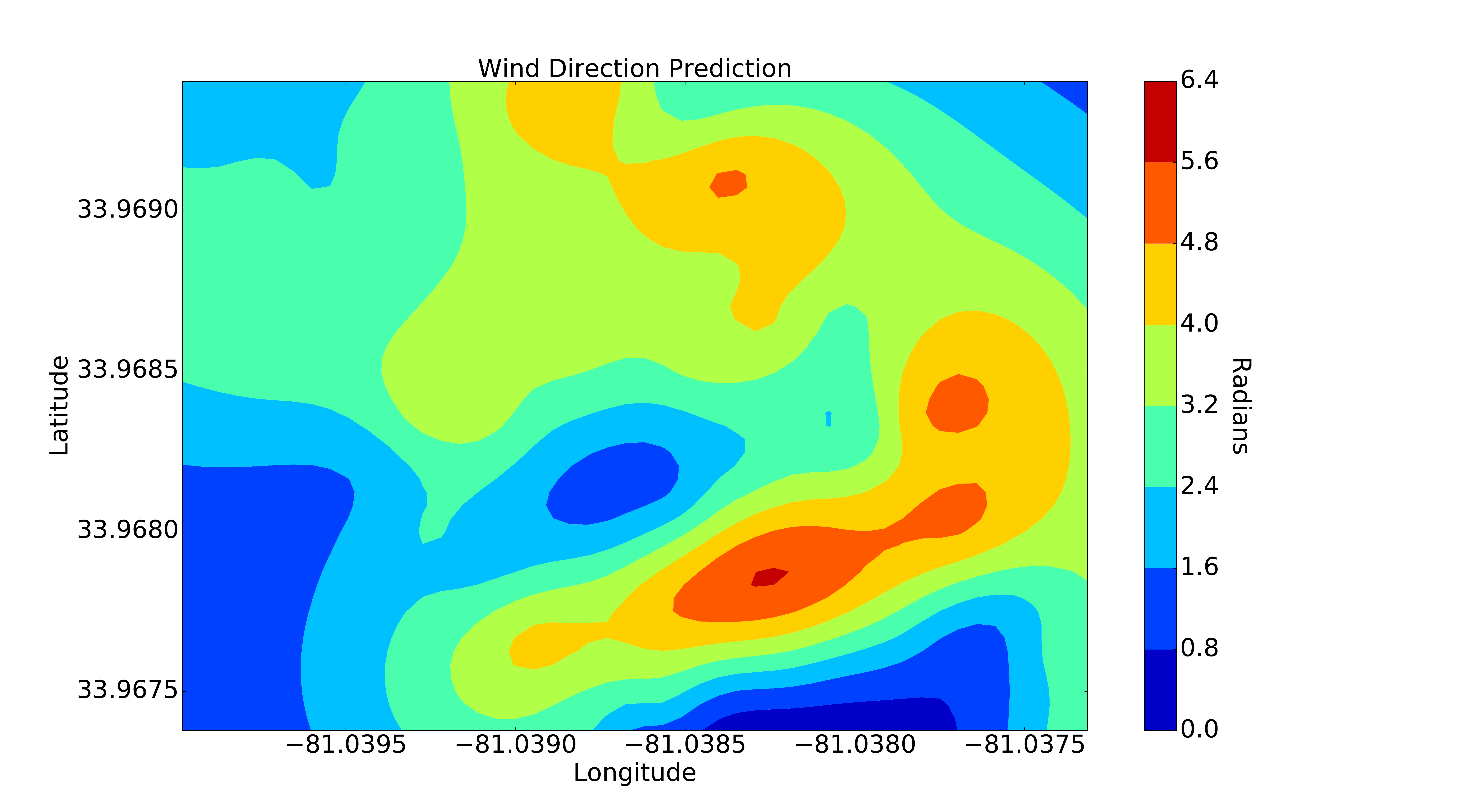}}\vspace{-0.15in}
		\end{tabular}
			\caption{(a) Wind Speed Prediction Map for Congaree River, SC.  (b) Wind Direction Prediction Map for Congaree River, SC. 
			\label{fig:predictions}}
	\end{figure*}

\begin{figure*}[t]
		\centering
		\leavevmode
		\begin{tabular}{cc}
		\subfigure[\label{fig:current_speed_pred}] {\includegraphics[width=0.54\textwidth, trim={2cm 0 0 2.5cm},clip]{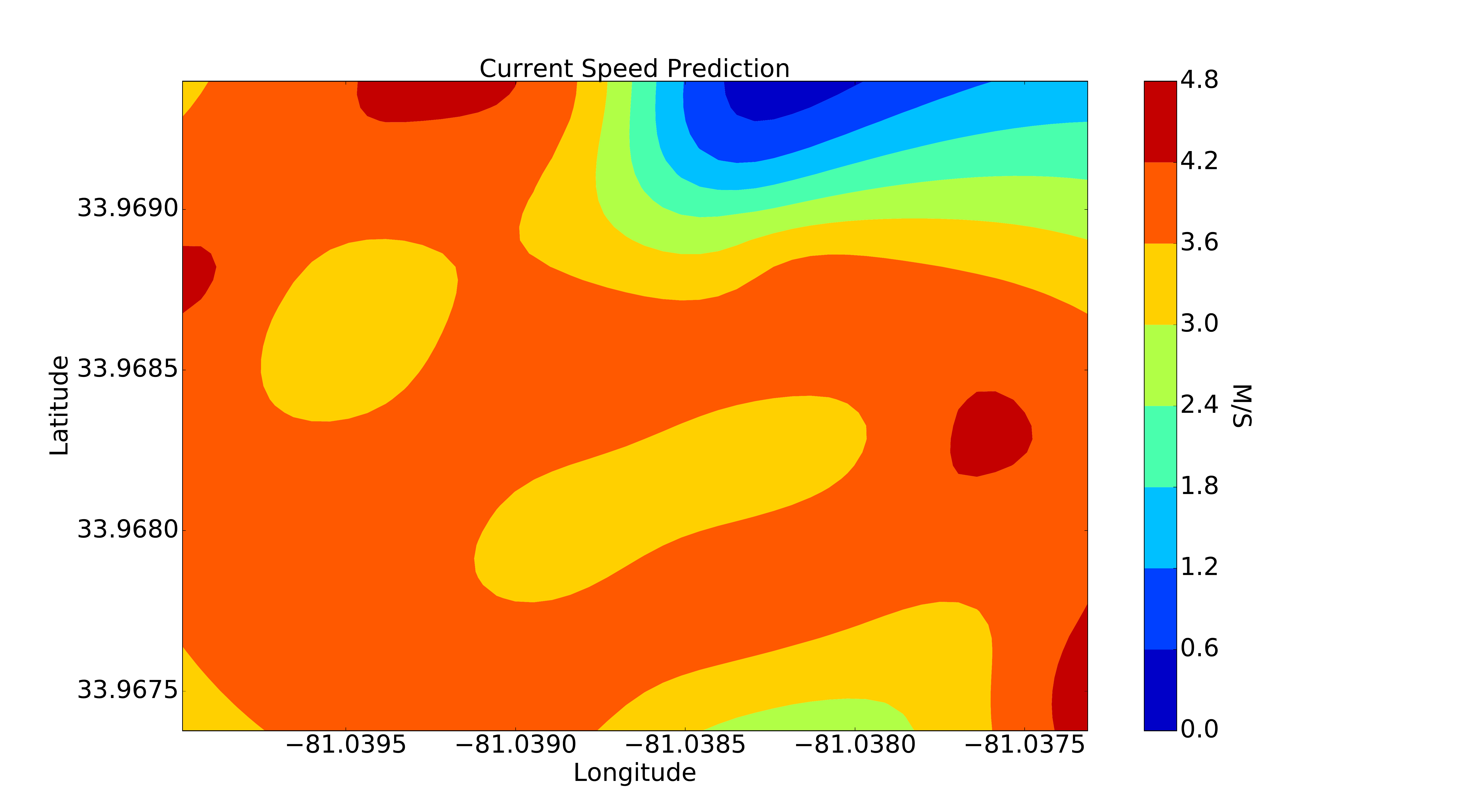}}&
			\subfigure[\label{fig:current_dir_pred}] {\includegraphics[width=0.54\textwidth, trim={2cm 0 0 2.5cm},clip]{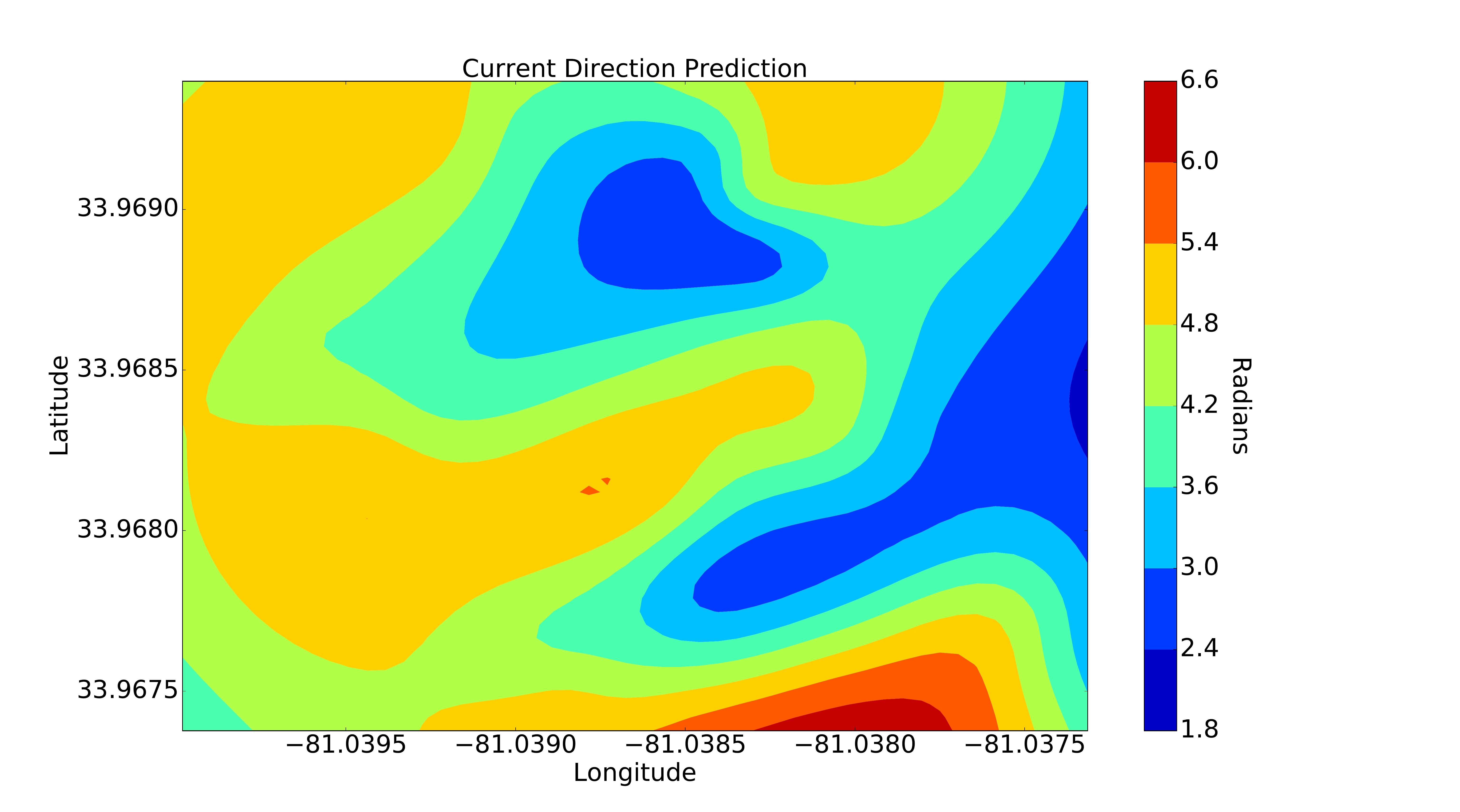}}\vspace{-0.15in}
		\end{tabular}
			\caption{(a) Current Speed Prediction Map for Congaree River, SC.  (b) Current Direction Prediction Map for Congaree River, SC. 
			\label{fig:current_predictions}}
	\end{figure*}
Some preliminary experiments were completed to fine tune the PID controller parameters and to assess the effects of taking the ASV, whose controller was tuned in the lake, to the river and evaluating its performance. As shown in Figure \ref{fig:grid_search_windy}, clearly the ASV was not able to maintain the planned trajectory.

The main set of experiments included data collection under different conditions. In particular, first, we collected wind and current measurements by tying the ASV so that is stationary. In this way, a baseline is available to compare with, when data is collected as the ASV is moving. 
Second, we planned different waypoint missions characterized by different patterns -- i.e., parallel and, perpendicular to current patterns -- to collect data from different orientations. 
\invis{In all experiments the ASV is configured, tuned and weighted to the same specifications. The parameters tuned on the ASV are the PID controller coefficients for the steering servo, and the PID coefficients on the throttle servo. Steering controller intuitively controls yaw tracking while the throttle PID controller is mapped to control acceleration throttle and cruise throttle.}
\invis{As we further our research and increase the robustness and reliability of the Jetyak fleet, we are now pushing the boundaries and deploying in unstable environments. }
\invis{\subsection{Focus on Wind in Lake Environments}}

\begin{figure*}[b]
		\centering
		\leavevmode
		\begin{tabular}{cc}
		\subfigure[\label{fig:actual_wind}] {\includegraphics[width=0.5\textwidth, trim={5cm 0 0 2.5cm},clip]{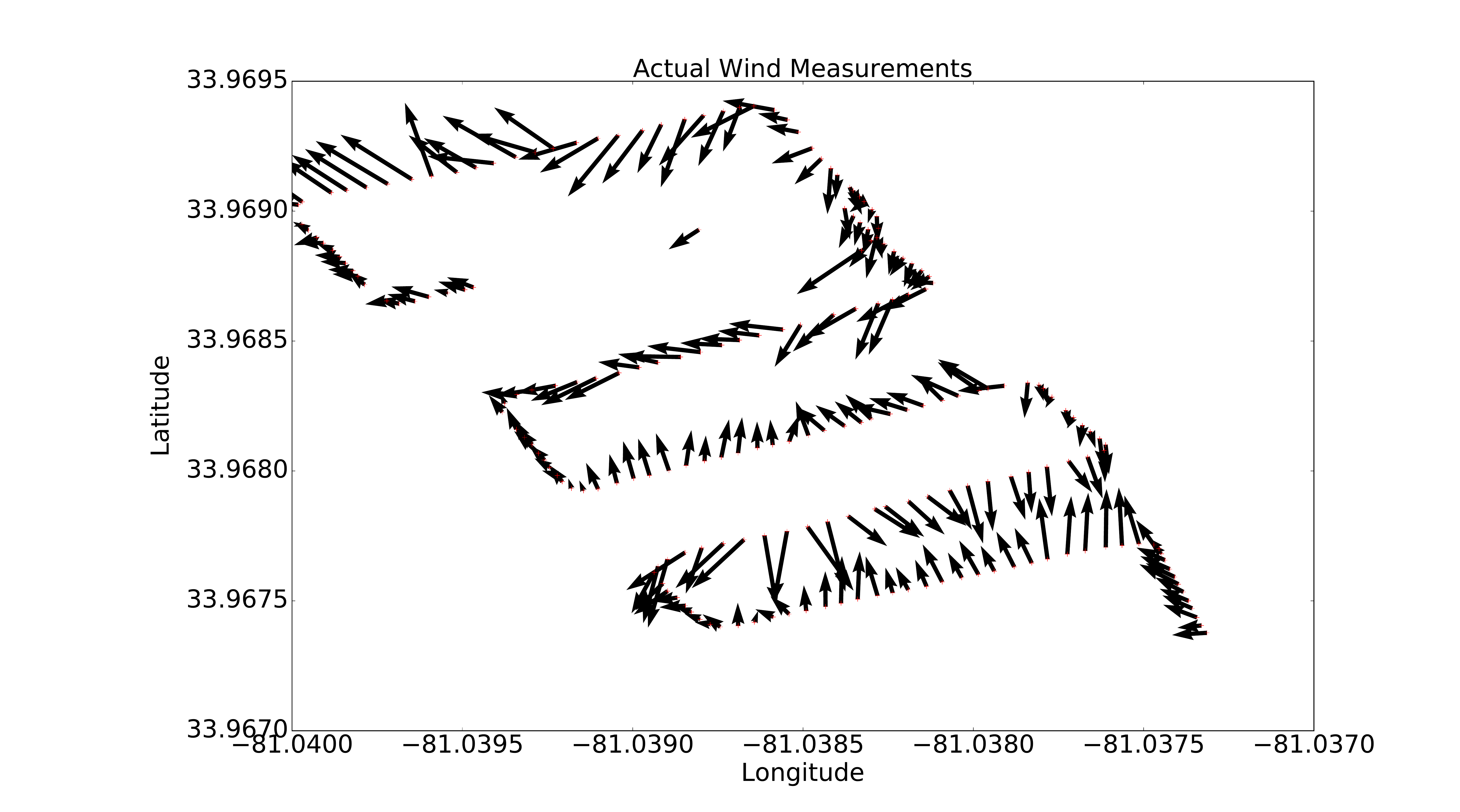}\label{actual_wind}}&
			\subfigure[\label{fig:actual_current}] {\includegraphics[width=0.5\textwidth, trim={5cm 0 0 2.5cm},clip]{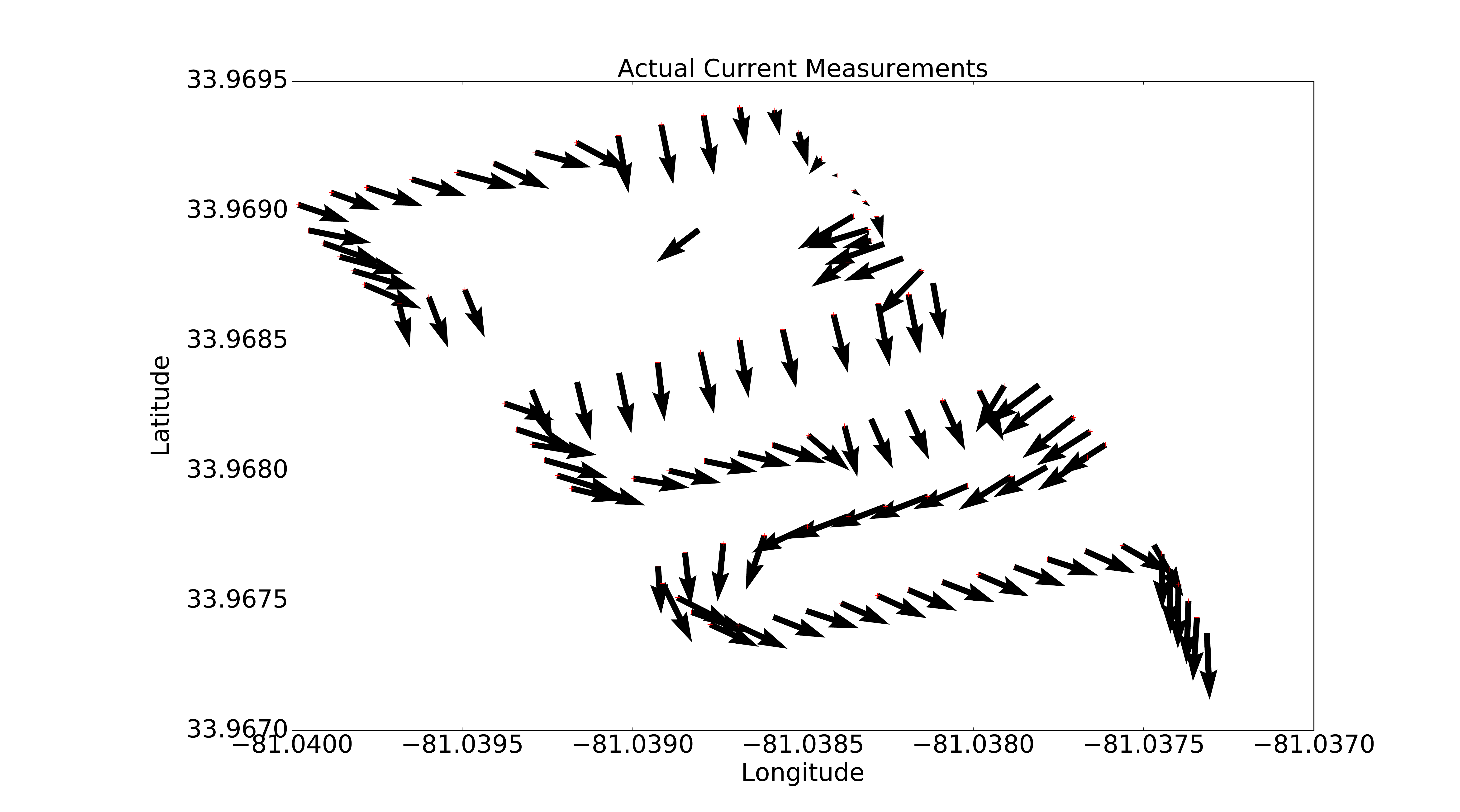}}
		\end{tabular}
			\caption{Field trial on the Congaree River, SC on the trajectory in Figure  \ref{fig:river_perp}. (a) Recorded wind measurements. (b) Recorded current measurements. 
			\label{fig:congaree_data}}
\end{figure*}

Initial collected data in Figure \ref{fig:depth_map} reflects the resulting topographical map of the riverbed after data collection using the coverage path in Figure \ref{fig:river_perp}. Wind and current data collected from the Congaree River clearly characterizes highly dynamic currents, changing winds, and a highly variant depth -- see Figure \ref{fig:congaree_data} for recorded wind and current measurements; and Figures \ref{fig:predictions} and \ref{fig:current_predictions} for the predicted wind speed and direction, and current speed and direction, respectively, over the region. Figure \ref{fig:actual_wind} illustrates successful data collection and modeling of wind forces experienced in moderately windy conditions. Our system recorded sustained winds of \SI{8.5}{\m/\s} and gusts reaching \SI{12.5}{\m/\s}, accurate measurements according to the local weather recording station, which recorded sustained winds of \SI{9}{\m/\s} and gusts at \SI{14}{\m/\s} during the 3 hour trial. Given the mostly open terrain and the effects of the shoreline on wind patterns, our measurements are in-line with the recording station.  

\invis{\subsection{Dynamic Environments}}
As illustrated in Figure \ref{fig:congaree_data}, the average actual current measured while the ASV was navigating the cross sections of the river was \SI{5.0}{\m/\s}, compared to the docked stationary current measurement of \SI{3.0}{\m/\s}. The challenge with surface currents is establishing the ground truth due to the high degree of fluctuation with them. In addition to the previously mentioned lab calibrations, we ensure measurements are in the correct order of magnitude at the beginning of each field trial by recording the free floating GPS velocity. In this particular field trial, we approximated the ground truth current speed to be \SI{2.5}{\m/\s}. Due to the hull displacement, as expected, we recorded a slightly lower velocity using this free float method.

\invis{this frontier-type environment will allow us to gain better perspective to measure and model the dynamics in a volatile environment.}
\invis{Effective modeling along with testing our ASV in this environment will ensure a robust system that is accompanied with well-formed specifications and capabilities.}
\invis{Progressively, we deploy to more challenging settings. Our present location for testing, gathering and modeling environmental dynamics -- see Figure \ref{fig:river} --presents us with highly dynamic currents, changing winds, and a greater variance of depth.}

\vspace{-0.1in}
\section{Discussion}
Our technical and experimental contributions in this paper have produced a  baseline of inexpensive tools and methods for recording the external forces -- wind and current -- acting on an ASV. GP regression modeling allowed us to predict for a large area with sparse measurements collected during the experiments, as shown in Figures \ref{fig:predictions} and \ref{fig:current_predictions}. From the beginning of this endeavor, calibration of the analog current sensors to accurately reflect surface current required continuous refinement. Lab configuration and testing provided us an initial guideline on the best sensor configuration for a static vessel. \invis{However, the calibration of the sensors on a moving ASV proved to be most challenging.} Further on\hyp free-float and static ASV testing confirmed our measurements to be reasonable for the conditions. Currently, this system is stable for recording and observing nature's phenomena in action.

Of particular interest in these experiments is the confirmation of the close correlation between the depth (Figure \ref{fig:depth_map}) and the affected current patterns in Figure \ref{fig:current_predictions}. Verifying that our inexpensive system is able to collect and confirm US Geological Survey studies \cite{usgs}, encourages extending our research. As such, to improve the precision of predictions, our current work involves studying and analyzing different GP kernels to better emulate the short temporal livelihood of the predictions. The long-term goal is to combine all three models -- i.e., current, depth, wind -- into a comprehensive impact model that can be used to improve coverage and search algorithms and develop proactive controls to enable ASV operation in highly dynamic environments.

\invis{In addition, we plan to perform more field experiments and collect more wind, current, and depth data under different conditions, following the planned water releases by the electric company that controls the upstream hydro plant. Such experiments will be coupled with some inexpensive ways to record the current, one such method includes deploying streamlined floaters, so that the collected data by the ASV can be validated.
The final version of the paper will include  the above results.}
\section*{ACKNOWLEDGMENT}
The authors would like to thank the National Science Foundation for its support (NSF 1513203). The authors are grateful to Professor Enrica Viparelli from the Civil and Environmental Engineering Department for her assistance in ground truth measurements at the Hydraulics Laboratory.

\vspace{-0.1in}
\bibliographystyle{template/splncs_srt}
\bibliography{IEEEabrv,refs}

\end{document}